\documentclass[journal]{IEEEtran}
\ifCLASSINFOpdf
\else
\fi

\usepackage{cite}
\usepackage{amsmath,amssymb,amsfonts}
\usepackage{algorithmic}
\usepackage{graphicx}
\usepackage{textcomp}
\usepackage{wrapfig}
\usepackage{booktabs}
\usepackage{subcaption}
\usepackage{multirow}
\usepackage{makecell}
\usepackage{stfloats}
\usepackage{threeparttable}

\begin{document}
%
\title{Physically Plausible Data Augmentations for Wearable IMU-based Human Activity Recognition Using Physics Simulation}
%
%
%

\author{Nobuyuki Oishi, Philip Birch, Daniel Roggen, and Paula Lago
\thanks{This work received support from the EU H2020-ICT-2019-3 project ``HumanE AI Net'' (grant agreement number 952 026) and Huawei Technologies. Also, this research was enabled in part by support provided by Calcul Quebec and the Digital Research Alliance of Canada (alliancecan.ca).}
\thanks{N. Oishi, P. Birch, D. Roggen are with the School of Engineering and Informatics, University of Sussex, Brighton, East Sussex, UK (e-mail: n.oishi@sussex.ac.uk, p.m.birch@sussex.ac.uk, daniel.roggen@ieee.org).}
\thanks{P. Lago is with 
the Electrical and Computer Engineering Department, Concordia University, Montreal, QC, Canada (e-mail: paula.lago@concordia.ca).}}

%
%

\markboth{}%
{Oishi \MakeLowercase{\textit{et al.}}: Physically Plausible Data Augmentations for Wearable IMU-based HAR Using Physics Simulation}
%



\maketitle

\begin{abstract}
The scarcity of high-quality labeled data in sensor-based Human Activity Recognition (HAR) hinders model performance and limits generalization across real-world scenarios. Data augmentation is a key strategy to mitigate this issue by enhancing the diversity of training datasets. Signal Transformation-based Data Augmentation (STDA) techniques have been widely used in HAR. However, these methods are often physically implausible, potentially resulting in augmented data that fails to preserve the original meaning of the activity labels. In this study,  we introduce and systematically characterize Physically Plausible Data Augmentation (PPDA) enabled by physics simulation. PPDA leverages human body movement data from motion capture or video-based pose estimation and incorporates various realistic variabilities through physics simulation, including modifying body movements, sensor placements, and varying hardware-related effects. We compare the performance of PPDA methods with traditional STDA methods on three public datasets of activities of daily living and fitness workouts: REALDISP (34 classes), REALWORLD (8 classes), and MM-Fit (11 classes). First, we evaluate each augmentation method individually, directly comparing PPDAs to their STDA counterparts. Next, we assess how combining multiple PPDAs can reduce the need for initial data collection by varying the number of subjects used for training. Our experiments demonstrate consistent benefits of PPDAs, improving macro F1 scores by an average of 3.7 pp, with a maximum improvement of 13 pp, and achieving competitive performance with up to 60\% reduction in the number of subjects for training, compared to STDAs. As the first systematic study of PPDA in sensor-based HAR, these results highlight the advantages of pursuing physical plausibility in data augmentation and the potential of physics simulation for generating synthetic Inertial Measurement Unit (IMU) data for training deep learning HAR models. This cost-effective and scalable approach therefore helps address the annotation scarcity challenge in HAR.
\end{abstract}

\begin{IEEEkeywords}
Data augmentation, human activity recognition, inertial measurement unit, physics simulation, deep learning, wearable computing
\end{IEEEkeywords}

%
\IEEEpeerreviewmaketitle

\section{Introduction}
%
%
%
%
\IEEEPARstart{T}{he}  advancement of wearable and mobile technologies has enabled motion sensor-based Human Activity Recognition (HAR), where activities are inferred from measurements obtained by triaxial accelerometers, gyroscopes, magnetometers, or their combination in Inertial Measurement Units (IMUs). This has led to diverse applications such as healthcare, fitness monitoring, and entertainment \cite{lara2012survey, liu2021overview}. Despite its significant potential, developing sensor-based HAR systems faces a major challenge: the high cost of collecting high-quality labeled data for training machine learning models. This issue is particularly pressing for deep learning-based methods, which require large, diverse datasets to achieve high performance and generalizability \cite{plotz2018deep, nweke2018deep}.

Datasets for training HAR models must account for various factors, including differences in how individuals perform activities, diverse environmental conditions, and variations in sensor hardware and placement \cite{stisen2015smart, banos2014dealing, roggen10_pervasive}. However, collecting data that captures all such variabilities is often impractical due to high costs, making it difficult to develop models that generalize well to real-world settings. To address this, data augmentation is often employed \cite{chen2021deep}. By synthetically increasing the diversity of training datasets, data augmentation enhances model robustness to these inherent variabilities and reduces the need for extensive data collection efforts.

Often, what we refer to as Signal Transformation-based Data Augmentation (STDA) methods for time-series data are employed \cite{um2017data, mohammad2023enhanced, jeong_sensor-data_2021, ohashi2017augmenting, wen2021time}. These methods vary the signal by applying simple signal processing methods such as adding noise, scaling, rotation, and time-warping. While these STDA approaches have been shown to enhance model performance to some extent, they often lead to physically implausible data augmentation (see Table \ref{tab:trans-aug-summary}). This lack of physical plausibility might introduce the risk of data-label mismatch, where the augmented data no longer represents the intended activity class.

In this paper, we introduce and systematically characterize Physically Plausible Data Augmentation (PPDA) to investigate its potential benefit on HAR model performance and its potential to reduce the need for extensive labeled data.

We implement PPDA methods using WIMUSim \cite{oishi7wimusim}, an open-source physics-based wearable IMU simulation framework, which models wearable IMU data using four key parameters: Body (skeletal model), Dynamics (movement patterns), Placement (device positioning), and Hardware (IMU characteristics). These parameters account for real-world variabilities affecting wearable IMU data. Using WIMUSim, we first identify these parameters through gradient descent-based optimization, leveraging real IMU data alongside concurrently collected body movement data from motion capture systems or video-based pose estimation techniques. This process ensures the high-fidelity simulated IMU data. Then, by adjusting these parameters, PPDA generates augmented IMU data with physically plausible variations, while maintaining high fidelity in the simulated IMU data.

We conduct experiments to compare PPDA methods with traditional STDA techniques. Our evaluations are two-fold:
\begin{enumerate}
\item \textbf{Individual Comparison of Augmentation Methods:} we compare PPDAs to their closest STDA counterparts, including magnitude scaling/warping, time scaling/warping, rotating, and jittering.
\item \textbf{Impact on Reducing Data Collection Needs:} we analyze how effectively PPDAs can reduce reliance on extensive initial data collection by combining multiple data augmentation methods to enhance variations in the training data.
\end{enumerate}
Through these experiments, we demonstrate the advantages of PPDA in improving model performance and its potential to reduce the data requirements for training HAR models.

The primary contributions of this work are as follows:
\begin{itemize}
\item \textbf{Review of Data Augmentation and Virtual IMU Simulation in Wearable IMU-based HAR:} we review existing data augmentation techniques and the development of virtual IMU simulation in wearable IMU-based HAR, highlighting the emerging opportunity to utilize virtual IMU simulation for data augmentation. (Section \ref{sec:relatedwork})
\item \textbf{Development of Physically Plausible Data Augmentation Methods:} we detail the implementation of PPDA methods using WIMUSim and explain how it achieves better physical plausibility compared to traditional approaches (Section \ref{sec:method}).
\item \textbf{Evaluation of the Benefit of Physically Plausible Data Augmentation:} we evaluate individual and combined PPDA methods, analyzing their performance improvements (Section \ref{sec:individual-comparison}) and effectiveness in reducing reliance on extensive initial data collection (Section \ref{sec:reduce-data}).
\item \textbf{Code Release for Further Development:} we release our PPDA implementation as an open-source code base, enabling further research and development in wearable IMU-based HAR (https://github.com/XXX)\footnote{The code will be released at a public repository upon acceptance.}.
\end{itemize}

\section{Related Work}
\label{sec:relatedwork}
Deep learning models have achieved strong performance in HAR, but annotation scarcity—the lack of large and diverse datasets—remains a key bottleneck to improving model generalization across real-world scenarios \cite{chen2021deep}.
Variability in wearable IMU data arises from multiple factors, including variations in human movement and sensor placement, as well as hardware-related effects such as signal noise and bias.
Since deep learning models require large, diverse datasets for better generalization, data augmentation plays a critical role in enriching training data to compensate for the limitations of real-world data collection.

We first review signal transformation-based data augmentations, the most widely used data augmentation methods in sensor-based HAR, which serve as the basis for our evaluation. We then discuss recent advancements in data augmentation that utilize virtual IMU simulation as an emerging approach, as our method leverages physics-based simulation to introduce physically plausible transformations. Finally, for completeness, we mention signal generation-based data augmentation methods, but as they follow a fundamentally different approach and are not closely related to this study, we keep this brief.

\subsection{Signal Transformation-based Data Augmentation}
Signal Transformation-based Data Augmentations (STDAs) include signal transformation techniques such as magnitude scaling, magnitude warping, time scaling, time warping, rotating, and jittering. Table \ref{tab:trans-aug-summary} summarizes these methods, highlighting their associated variations and limitations. These methods apply simple mathematical transformations to IMU signals, making them computationally efficient and easy to integrate into training pipelines. While some of these STDAs originate from other time series domains such as audio processing (e.g., amplitude scaling), their application to activity recognition does not guarantee the preservation of physical plausibility. Others are superficially informed by IMU signal characteristics but remain primarily designed for simplicity and ease of implementation, rather than accurately modeling real-world sensor signal variations. As a result, STDAs often lead to physically implausible IMU signal augmentations.

\begin{table*}[]
\scriptsize
\caption{Summary of common signal transformation-based data augmentation methods used in sensor-based HAR.}
\label{tab:trans-aug-summary}
\begin{tabular}{p{2cm}p{9.1cm}p{5.7cm}}
\toprule
\textbf{Method} & \textbf{Description} & \textbf{Associated Real-World Variation \& Limitation} \\
\midrule
Magnitude Scaling\newline \cite{um2017data, mohammad2023enhanced, kalouris2019improving, jeong_sensor-data_2021, zhou2024autoaughar} & Uniformly scale IMU signal \( X \in \mathbb{R}^{T \times C} \) by a factor \( \alpha \), where \( T \) is the number of time steps and \( C \) is the number of sensor channels. \( \alpha \) is sampled from a normal distribution \( N(1, \sigma^2) \) to increase or decrease the signal amplitude. The transformation is applied as: $X' = \alpha X$  & \multirow{2}{5.6cm}{Associated Variation: Changes in movement amplitude (within the same time frame). \\ 
Limitation: Fails to capture the effects of multi-joint kinematics in human movements and scales the entire accelerometer and gyroscope signals. For the accelerometer, this means scaling the gravitational component of acceleration readings as well, which should remain constant.}  \\  \addlinespace[4pt] \cline{1-2} \addlinespace[2pt]
Magnitude Warping \cite{um2017data, zhou2024autoaughar} & Scale IMU signal \( X \) by a time-varying scaling factor \( \boldsymbol{\alpha} \in \mathbb{R}^{T} \), where \( \boldsymbol{\alpha} \) is generated by sampling values at \( K \) knots from a Gaussian distribution \( N(1, \sigma^2) \) and interpolating them with a cubic spline. The transformation is applied element-wise as: $X' = \boldsymbol{\alpha} \odot X $  &   \\ \midrule
Time Scaling~\cite{xia_virtual_2022, zhou2024autoaughar}  & Uniformly stretch or compress IMU signal \( X \) along the time axis by a factor \( \beta \), modifying the time index as \( t' = t / \beta \). The resampled signal is obtained via interpolation: $X' = \text{interp}(X, t')$ where \( \text{interp}(X, t') \) denotes an interpolation function.
 & \multirow{2}{5.6cm}{Associated Variation: Changes in movement speed. \\ Limitation: Fails to capture the physically linked variations in acceleration and angular velocity that occur with changes in movement speed, as these methods modify only the temporal aspect of the signals while leaving the magnitude unchanged.} \\ \addlinespace[0pt] \cline{1-2} \addlinespace[4pt]
Time Warping \cite{jeong_sensor-data_2021, um2017data, zhou2024autoaughar}  & Modify IMU signal \( X \) using a non-uniform time warping function \( w(t) \), which controls local speed variations. The transformed time indices are computed as \( t' = w(t) \), where \( w(t) \) is generated via cubic spline interpolation over \( K \) knots. The resampled signal is obtained via interpolation: $X' = \text{interp}(X, w(t))$ &  \\ \midrule
Rotating \cite{um2017data, ohashi2017augmenting, kalouris2019improving, mohammad2023enhanced, jeong_sensor-data_2021}   & Rotate IMU signal \( X \in \mathbb{R}^{T \times N \times 3} \), which contains \( N \) triaxial sensor signals (e.g., acceleration, gyroscope), using a randomly generated 3D rotation matrix \( R \in \mathbb{R}^{3 \times 3} \). The transformed signal is computed as: $ X'_{n} = R X_{n}$, where $n$ indexes each triaxial sensor. The same rotation matrix $R$ is applied uniformly across all time steps. & Associated Variation: Changes in sensor orientation. \newline Limitation: Often applies an unrealistic range of rotations and does not account for sensor positioning shifts (e.g., displacements due to wearing preferences.) \\ \midrule
Jittering  \cite{mohammad2023enhanced, kalouris2019improving, um2017data, jeong_sensor-data_2021, ohashi2017augmenting, zhou2024autoaughar} & Perturb IMU signal \( X \) by adding Gaussian noise \( \eta \) sampled from \( N(0, \sigma^2) \) to each time step and sensor channel. The transformed signal is computed as: $X' = X + \eta, \quad \eta \sim N(0, \sigma^2)$ & Associated Variation: Differences in IMU noise levels. \newline Limitation: May not accurately approximate real noise variations, as it adds synthetic noise to existing sensor noise and does not account for other hardware-related effects such as sensor bias (constant offset). \\
\bottomrule
\end{tabular}
\end{table*}

Ohashi et al. \cite{ohashi2017augmenting} explored the impact of physically constrained augmentation, focusing on the rotation of a wrist-worn device equipped with 9-axis IMU and EMG sensors. By limiting rotations to be applied to directions that are physically plausible, i.e. only around the wrist, they demonstrated improved performance, 2 to 5 percentage points (pp), compared to augmentations without such constraints on a dataset for a 3-class action classification task (holding, twisting, other). This suggests the benefit of integrating physical constraints in data augmentation to ensure that augmented data adheres to the physical principles underlying sensor-based HAR. However, their study was limited to rotating augmentation and was not evaluated on other common STDA methods. Additionally, its findings were demonstrated on a single dataset, making it unclear whether the benefits of physical plausibility generalize across different datasets and activity recognition tasks.

\subsection{Emerging Opportunity: Virtual IMU Augmentation}
The advancement of virtual IMU simulation presents a new avenue for generating IMU signals from datasets acquired through different modalities, such as motion capture or video-based pose estimation. Virtual IMU simulation refers to the process of generating simulated IMU data, akin to what would be produced by physical IMUs placed on a moving human body. This is often achieved by applying geometrical transformations to 3D human motion data, allowing sensor signals to be computed based on body kinematics. Prior studies have explored virtual IMU simulation from motion capture \cite{young2011imusim, lago2019measured, pellatt2021mapping} or video-based pose estimation \cite{kwon2020imutube, kwon2021approaching, santhalingam2023synthetic}, but these efforts have primarily focused on creating virtual IMU datasets from pre-existing video or motion capture data rather than data augmentation for sensor-based HAR.

One of the earliest attempts to apply virtual IMU simulation for data augmentation was made by Xia and Sugiura \cite{xia_virtual_2022}. They proposed a simulation-based augmentation method using a Unity-based virtual IMU sensor model with spring joints. This model introduces variations, particularly in acceleration in the vertical direction of the body, to mimic accelerometer readings generated during dynamic movements, such as those observed in aerobic exercises. However, their approach required manual feature crafting and dimensionality reduction to mitigate the domain gap between virtual and real IMU data and was tested with classical machine learning models such as SVMs and random forests, leaving the potential benefit for deep learning unexplored. Additionally, their evaluation was restricted to a small dataset consisting of only three exercise activities (reverse lunge, warm-up, and high knee tap), leaving its generalizability to broader movement contexts uncertain.

Leng et al. \cite{leng2023generating} proposed to integrate a generative model for 3D human motion synthesis with virtual IMU simulation. Specifically, they utilize T2M-GPT \cite{zhang2023t2m}, a model that generates human motion sequences from textual descriptions, to create synthetic movement data, from which virtual IMU signals are derived. Although this approach differs from our focus, as it generates entirely new motion sequences rather than transforming existing data for augmentation, their work suggests an interesting application of virtual IMU simulation.

While these approaches demonstrate the potential of virtual IMU simulation for data augmentation, these simulations suffer from a domain gap between virtual and real IMU data. Models trained exclusively on virtual IMU data, particularly deep learning-based models, often perform significantly worse than those trained on real IMU data \cite{kwon2020imutube, leng2023generating}.

WIMUSim \cite{oishi7wimusim} mitigates the domain gap between virtual and real IMU data—largely caused by fidelity limitations in simulated signals—through an additional system identification step, which aligns virtual IMU data with real-world observations. 
It models 6-axis IMU data—accelerometer and gyroscope signals—using four parameter sets: Body ($B$), which represents a skeletal model of the human body; Dynamics ($D$), which describes joint orientations and global translation over time; Placement ($P$), which specifies sensor positioning relative to body segments; and Hardware ($H$), which characterizes sensor noise and bias.
Using these parameters, WIMUSim analytically computes accelerometer and gyroscope signals that remain physically consistent with the underlying body movements, sensor placement, and hardware-related effects.
Through an optimization process, these parameters ($B, D, P, H$) are refined based on real IMU data and motion information captured via video-based or other motion tracking systems, ensuring that the simulated signals closely resemble real IMU measurements.
As a result, WIMUSim generates high-fidelity virtual IMU data that effectively bridges the gap between synthetic and real-world sensor signals, achieving comparable performance when used to train deep learning models \cite{oishi7wimusim}.

\subsection{Signal Generation-based Data Augmentation}
Another distinct class of data augmentation methods is signal generation-based data augmentations using generative models, such as Generative Adversarial Networks (GANs) \cite{i_activitygan_2020, ramachandra_transformer_2021, chen2021gan, ditthapron2023adl, gerych2021gan}. These generation-based methods learn complex representations of data and synthesize new samples for enriching HAR datasets. For instance, ADL-GAN \cite{ditthapron2023adl} utilizes GANs' style translation capabilities \cite{zhu2017cyclegan, choi2018stargan}, such as transforming jogging data into walking data and adapting data from one subject to emulate that of another.

However, generation-based approaches usually lack explicit mechanisms to ensure physically plausible IMU readings. Moreover, Zhou et al. demonstrated that STDAs with optimized data augmentation policies can outperform signal generation-based approaches in sensor-based HAR tasks \cite{zhou2024autoaughar}. Beyond these limitations, a key challenge of generative models is their reliance on large, diverse real datasets for effective training. This can be particularly problematic in wearable sensor-based HAR, where new sensor types, combinations, and device variations continue to emerge, making it difficult to acquire sufficiently diverse datasets in the first place.

In contrast, our PPDA approach leverages the flexibility of physics simulation to integrate domain knowledge in designing physically meaningful augmentations, without requiring large-scale real IMU datasets for training. This makes PPDA particularly advantageous in scenarios where data collection is expensive or limited. Given these fundamental differences, data augmentation using generative models serves a different purpose and is not directly comparable to our approach.

\subsection{Summary and Research Gap}
STDAs have been widely used in sensor-based HAR, yet they often yield physically implausible augmented data (as explained in Table \ref{tab:trans-aug-summary}). Meanwhile, virtual IMU simulation offers a promising approach to achieving physically plausible data augmentation. Although previous studies have explored introducing physical constraints in data augmentation or simulation-based data augmentation, these efforts remain limited in scope and scale. In particular:

1) The benefits of physically plausible data augmentation remain insufficiently explored. While prior work~\cite{ohashi2017augmenting} suggested its potential advantage, evaluations were largely focused on rotation constraints with a small dataset comprising only 3 classes, leaving the impact of physical plausibility on other augmentation methods and broader activity sets unclear.

2) The application of virtual IMU simulation for data augmentation remains underexplored. While virtual IMU simulation has been used to generate synthetic datasets, its effectiveness in augmenting existing sensor-based HAR datasets has not been systematically studied. In addition, a key limitation lies in the lower fidelity of virtual IMU data, which contributes to the sim-to-real domain gap and may hinder the effectiveness of physics-based simulation for real-world HAR models.

To systematically investigate these limitations, we leverage WIMUSim~\cite{oishi7wimusim}, which provides high-fidelity virtual IMU data suitable for training deep learning models. It also enables the introduction of controlled variations through modifications of its key parameters—Body, Dynamics, Placement, and Hardware—forming the basis for our PPDA methods. We evaluate these methods using three public HAR datasets, covering 8 to 34 classes, to examine their effectiveness across a broad range of activities.

\section{Physically Plausible Data Augmentations}
\label{sec:method}
In this section, we describe how we design PPDAs using WIMUSim. By modifying its parameters---Dynamics ($D$), Placement ($P$), and Hardware ($H$)---, we introduce physically plausible variations into 6-axis IMU data (accelerometer and gyroscope signals) while preserving realistic motion dynamics, sensor placement, and hardware-related effects. This approach enables the generation of diverse, high-fidelity virtual IMU data suitable for training deep learning models.

\begin{figure*}[t]
  \centering

  \hfill
  \begin{subfigure}[b]{1.0\textwidth}
    \centering
    \includegraphics[width=\textwidth]{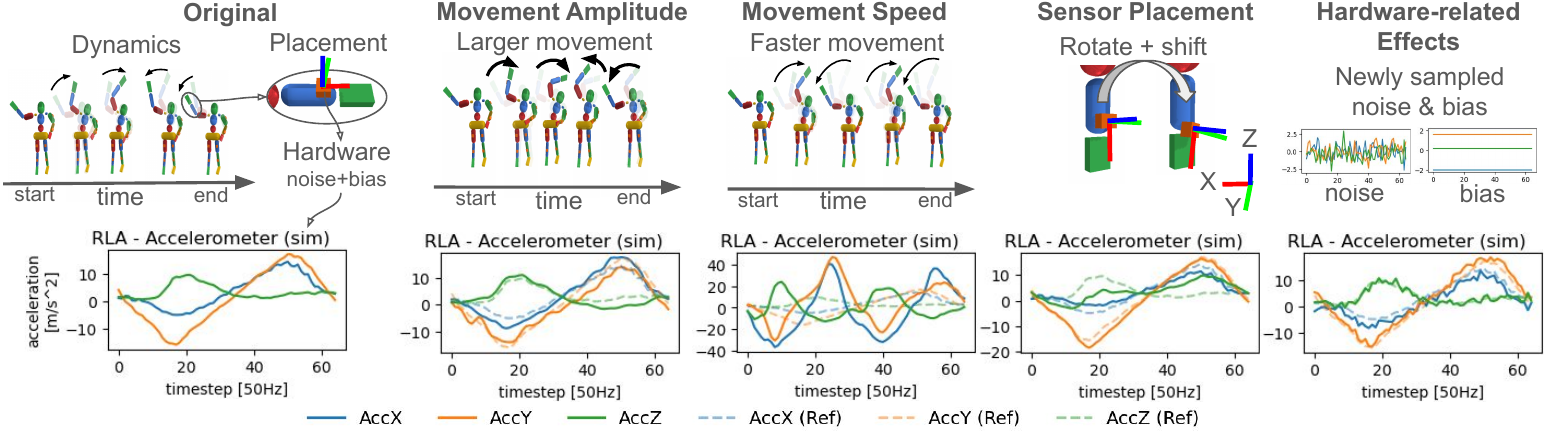}
    \caption{PPDAs: Physically Plausible Data Augmentations}
    \label{fig:ppda}
  \end{subfigure}
  \hfill
  \vspace{0pt}
  \begin{subfigure}[b]{1.0\textwidth}
    \centering
    \includegraphics[width=\textwidth]{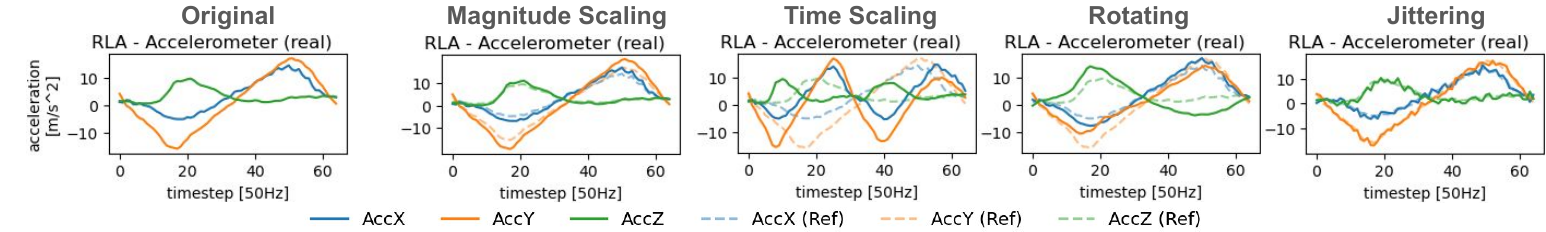}
    \caption{STDAs: Signal Transformation-based Data Augmentations}
    \label{fig:stda}
  \end{subfigure}
  \caption{(a) Illustration of how PPDAs introduce variations to IMU data using WIMUSim. Movement amplitude and speed are modified via the Dynamics parameter, while sensor placement and hardware-related effects are controlled by the Placement and Hardware parameters, respectively. The accelerometer data shown is from an IMU (orange cube) on the right wrist during a lateral bend movement. 
 (b) Comparison of corresponding STDA methods, which apply signal transformations directly to real IMU data. For instance, to increase movement speed, the PPDA ``Movement Speed'' (top row, middle column) doubles the humanoid's movement speed, resulting in compressed signals along the time axis and increased acceleration. Specifically, the first peak of the reference Z-axis signal shifts from \( t = 20 \) to \( t = 10 \), and acceleration increases from approximately \(20\,\text{m/s}^2\) at \( t = 50 \) to \(40\,\text{m/s}^2\) at \( t = 25 \) on the Y-axis. In contrast, the closest corresponding STDA counterpart, ``Time Scaling'' (bottom row, middle column), compresses the signal only in the time domain without altering amplitude, making the augmentation physically incorrect.}
  \label{fig:stda-ppda-comparison}
\end{figure*}

We present PPDAs in four categories: movement amplitude, movement speed, sensor placement, and hardware-related effects. Fig.~\ref{fig:stda-ppda-comparison} provides a visual comparison of PPDAs and the most closely related STDAs, demonstrating how PPDAs introduce physically plausible variations across these four categories, whereas STDAs apply simple mathematical transformations, which often do not lead to physically plausible signals (as summarized in Table~\ref{tab:trans-aug-summary}). In the following subsections, we detail each category and explain how these improve physical plausibility compared to their STDA counterparts.

\subsection{Augmentation by Varying Movement Amplitude}
Movement amplitude refers to the range of joint motion during an activity. Individuals exhibit variations in movement amplitude due to differences in flexibility, strength, and intent. For example, a person may take longer strides while walking or bend their knees more deeply when squatting.

WIMUSim describes the evolution of a humanoid's movement over time through the \( D \) parameter. This parameter consists of a sequence of rotation quaternions \( D^{j} = \{q^j_1, q^j_2, \ldots, q^j_{T}\} \) for each joint \( j \), where each quaternion \( q^j_t \) at time \( t \) represents the joint’s orientation relative to its parent. These quaternions are derived from motion capture or video-based pose estimation. To simulate variations in movement amplitude in a physically plausible way, we modify the \( D \) parameter. 
A rotation quaternion \( q_t^j \) is expressed as:  
\[
q_t^j = \cos\left(\frac{\theta}{2}\right) + \sin\left(\frac{\theta}{2}\right)(u_x i + u_y j + u_z k),
\]
where \( \theta \) is the rotation angle, and \( \mathbf{u} = (u_x, u_y, u_z) \) is the unit vector along the axis of rotation. By applying magnitude scaling or magnitude warping to the sequence of rotation angles \( \theta_{1} \ldots \theta_{T} \), instead of directly modifying IMU signals, we adjust the amplitude of joint movement either uniformly or dynamically. For example, increasing the rotation angle of the knee joint would simulate a deeper squat, while decreasing the angle would simulate a shallower one. The resulting virtual IMU data---accelerometer and gyroscope signals---reflects these variations in movement amplitude while preserving natural motion dynamics and axes of joint rotations.

Fig.~\ref{fig:ppda} (Movement Amplitude) illustrates a human model performing a lateral bend, where the joint rotation angles are scaled by a factor of 1.25. This results in more pronounced bending of the arms compared to the original, increasing the movement amplitude. Unlike STDA methods, which uniformly scale all signal components, our PPDA implementation adjusts movement amplitude in a way that follows the natural motion dynamics without affecting the gravitational component of acceleration readings, resulting in more physically plausible augmented data.

\subsection{Augmentation by Varying Movement Speed}
Movement speed refers to how quickly an activity is performed. In everyday life, movement speed varies from person to person also based on context and intent, for example, cycling at different speeds, or lifting a weight explosively versus with slow control.

To introduce variations in movement speed, we modify the playback speed of WIMUSim by adjusting the \( D \) parameter over time. Instead of applying time scaling or time warping directly to IMU signals, we apply these operations to the \( D \) parameter—the temporal progression of joint orientations. This allows WIMUSim to simulate the same movement trajectory at a different speed—either uniformly or with local speed variations—while preserving realistic motion dynamics and resulting in physically consistent changes in acceleration and angular velocity.

Fig.~\ref{fig:ppda} (Movement Speed) illustrates a human model performing lateral bends at twice the original speed, hence completing two repetitions within the same duration. The corresponding acceleration data shows that each repetition takes half the original time, with increased peak accelerations.

In contrast, STDA time scaling and time warping only stretch or compress recorded sensor signals, without adjusting acceleration and angular velocity accordingly. Our PPDA approach modifies movement speed directly in the simulation environment, ensuring that virtual IMU data reflect these variations in a physically consistent manner, where slower/faster movements naturally lead to smaller/larger acceleration and angular velocity.

\subsection{Augmentation by Varying Sensor Placement}
Sensor placement refers to the position and orientation of a wearable IMU on the body. For example, a smartwatch may be worn on the dominant or non-dominant wrist, with the screen facing inward or outward, or positioned slightly higher or lower due to wrist thickness or body composition. Similarly, smart rings can be worn on different fingers or at varying orientations on the same finger.

In WIMUSim, sensor placement is controlled by the parameter \( P \), which defines the position and orientation of each virtual IMU relative to its associated body joint. For a given IMU \( u \) attached to joint \( j \), the placement is specified by a relative position vector \( P_{rp}^{(j, u)} \in \mathbb{R}^3 \) and a relative orientation quaternion \( P_{ro}^{(j, u)} \in \mathbb{H} \). These values can also be modified to simulate sensor placement variations.

Fig.~\ref{fig:ppda} (Sensor Placement) illustrates the modification of IMU placement on the right wrist and how it affects acceleration readings. Unlike STDA rotating (Fig.~\ref{fig:stda}), which randomly rotates IMU signals without considering the physical constraints of sensor attachment, our PPDA approach adjusts sensor placements in the simulation environment, ensuring both position and orientation adjustments remain realistic.

Moreover, the flexibility in our PPDA approach allows systematic exploration of realistic sensor placement variations by utilizing placement diversity derived from data collected across multiple subjects. By pairing the \( P \) parameter from one subject with other parameters \( B, D, H \) from another, we can simulate cross-subject variations in sensor placement, which is particularly valuable in multi-device setups. These capabilities ensure that models are trained on data representing a diverse and realistic range of sensor placements.

\subsection{Augmentation by Varying Hardware-related Effects}
IMU measurements are subject to hardware-related errors, including noise and bias, which arise from various sources in both accelerometers and gyroscopes. These errors vary depending on sensor quality and calibration, with consumer devices—particularly those equipped with low-cost sensors—exhibiting greater noise and bias due to manufacturing variations and less rigorous calibration compared to research-grade IMUs.
Noise primarily originates from electronic circuits (thermal noise) and vibrations, introducing small high-frequency fluctuations in sensor readings. Temperature variations further affect sensor stability by altering the physical and electrical properties of the sensing elements. Bias consists of an initial offset present at startup (turn-on bias) and a gradual drift over time (bias instability), both of which contribute to measurement inaccuracies.

In WIMUSim, the Hardware parameter \( H \) defines per-sensor noise and bias characteristics, with separate parameters for each accelerometer and gyroscope in every IMU. The standard deviation vector \( \boldsymbol{\sigma} \in \mathbb{R}^3 \) defines the noise level for each axis of a triaxial sensor, from which Gaussian noise is sampled, and the bias vector \( \boldsymbol{b} \in \mathbb{R}^3 \) specifies a constant offset per axis. These values are first identified during the parameter identification process, where WIMUSim isolates sensor-specific noise and bias components from real IMU data. To introduce hardware-related variations during augmentation, we vary \( \boldsymbol{\sigma} \) and \( \boldsymbol{b} \) for each IMU’s accelerometer and gyroscope: Gaussian noise is sampled at each timestep from \( N(0, \boldsymbol{\sigma}) \), while bias is modeled as a constant offset, sampled once per simulation run, then added to the simulated IMU signals.

In contrast, jittering—a widely used STDA technique for simulating hardware noise—simply adds additional Gaussian noise on top of existing sensor noise. This approach prevents the simulation of lower-noise scenarios and may result in excessively noisy data when applied to already noisy sensor readings. Moreover, it does not account for sensor bias, which is prevalent in real-world consumer devices \cite{stisen2015smart}.

Fig.~\ref{fig:ppda} (Hardware-related Effects) shows an example of sampled noise and bias and their effect on augmented accelerometer data in our PPDA approach. In contrast, Fig.~\ref{fig:stda} (Jittering) illustrates the result of STDA jittering, where only noise is added without accounting for sensor bias.

\section{Experiments and Results}
To evaluate the proposed PPDA methods, we conducted two experiments: the first compares individual PPDA methods against their STDA counterparts, while the second investigates how well PPDA can reduce the need for extensive data collection by evaluating its effectiveness when training data contain limited inter-subject variation.

\subsection{Datasets and Experimental Setup}
We used 6-axis IMU data (accelerometer and gyroscope) from three publicly available datasets—REALDISP, REALWORLD, and MM-Fit—covering activities of daily living and fitness workouts.  We selected these datasets for their availability of body motion data (e.g., IMU-based motion capture or video) required for parameter identification with WIMUSim, and their coverage of a range of activity types and sensor setups.

\textbf{REALDISP} \cite{banos2012benchmark} includes research grade IMU (XSens \cite{xsens2009}) data from 17 participants performing 33 fitness activities, plus a null activity class, with IMUs placed on the back, upper arms, lower arms, thighs, and shins. The dataset provides three placement configurations: 1) ideal placement, where IMUs are positioned by experts with consistent orientations; 2) self-placement, where participants attach IMUs themselves, introducing natural variability; and 3) mutual displacement, where IMUs are deliberately displaced by experts. We used the ideal and self-placement data in our experiments, including the null activity class during training and evaluation. With REALDISP, we define two evaluation scenarios to evaluate model generalization in different contexts in HAR: \textbf{1) user-independent scenario}: we use the dataset with ideal placement data and split it by subjects to evaluate generalization to unseen users. Subjects 1–10 are used for training, 11–12 for validation, and 13–17 for testing.
\textbf{2) sensor displacement scenario:} we use the ideal placement data from all 17 subjects for training, while we use self-placement data from subjects 14–17 for validation and from subjects 1–5, 7–12 for testing. Throughout the experiments, we focused on the Right Lower Arm (RLA) and Left Lower Arm IMUs as those are the typical placement of smartwatches or fitness trackers. IMU data were downsampled from 50 Hz to 25 Hz.

\textbf{REALWORLD} \cite{sztyler2016body} includes IMU and video recordings from 15 participants performing eight activities: walking, running, sitting, standing, lying, climbing stairs up, climbing stairs down, and jumping, without a null class. The dataset uses consumer devices for data collection: smartphones placed on the head, chest, waist, upper arm, thigh, and shin, and a smartwatch on the wrist. We split the dataset for a \textbf{user-independent scenario}, using data from subjects 1–5 and 8–12 for training, 6–7 for validation, and 13–15 for testing. Only a user-independent scenario was considered, as the dataset does not include multiple sensor placement configurations. IMU data were resampled to 30 Hz to match the RGB frame rate used for parameter identification.

\textbf{MM-Fit} \cite{stromback2020mm} comprises data from 10 participants performing 10 gym exercises, plus a null activity class, with IMU data collected from smartwatches worn on both wrists, an earbud worn on the left ear, and a smartphone in the right trouser pocket. The dataset contains 20 sessions, with some participants contributing more than one. We define a \textbf{user-independent scenario} and use one session per participant, including the null class, with recording IDs 0, 1, 2, 12, 13, and 16 for training, 17 and 19 for validation, and 18 and 20 for testing. For our experiments, we used data from the left earbud and both wrist IMUs, as these positions are common wearable sensor locations for fitness tracking. IMU data were resampled to 30 Hz for parameter identification, as with the REALWORLD dataset.

We trained two deep learning models: DeepConvLSTM \cite{ordonez2016deep} and AttendAndDiscriminate \cite{abedin2021attend}. DeepConvLSTM combines convolutional layers for local feature extraction with stacked Long Short-Term Memory (LSTM) layers to capture the temporal dynamics of features in sensor signals. AttendAndDiscriminate follows a similar architecture but introduces a channel interaction encoder between the convolutional and recurrent layers, which uses an attention mechanism to weigh the relative importance of different sensor channels. It also replaces the LSTM layers with GRUs.

\begin{table}[h]
\fontsize{7.25pt}{8.5pt}\selectfont
\centering
\begin{threeparttable}
\caption{Model configurations for each dataset.}
\label{tab:my-table}
\begin{tabular}{lccc}
\toprule
Dataset                                                                & REALDISP$^*$ & REALWORLD & MM-Fit  \\
\midrule
Window size/stride                                                     & 100/25   & 30/15     & 60/30   \\
Input channels                                                         & 12       & 42        & 18      \\
Output classes                                                         & 34       & 8         & 11      \\
DeepConvLSTM params        & 658,274  & 1,637,960 & 851,915 \\
AttendAndDiscriminate params & 522,916  & 1,256,842 & 667,405 \\
\bottomrule
\end{tabular}
\begin{tablenotes}
\item \parbox{\linewidth}{\vspace{0.5ex}$^*$The same model configuration was applied to both scenarios of the REALDISP dataset (user-independent and sensor-displacement).}
\end{tablenotes}
\end{threeparttable}
\end{table}

Table~\ref{tab:my-table} summarizes the dataset-specific model configurations, including window size/stride, number of input channels, output classes, and total trainable parameters. The size of batches was adjusted to maintain approximately 100 mini-batches per epoch: 256 (REALDISP, user-independent), 512 (REALDISP, sensor-displacement), 1024 (REALWORLD), and 128 (MM-Fit). All models were trained for 100 epochs, regardless of augmentation strategy, using the Adam optimizer with an initial learning rate of 0.001, decayed by a factor of 0.9 every 10 epochs. For each configuration, the best-performing model on the validation set was used for evaluation.

\begin{figure}[h]
  \centerline{\includegraphics[width=\columnwidth]{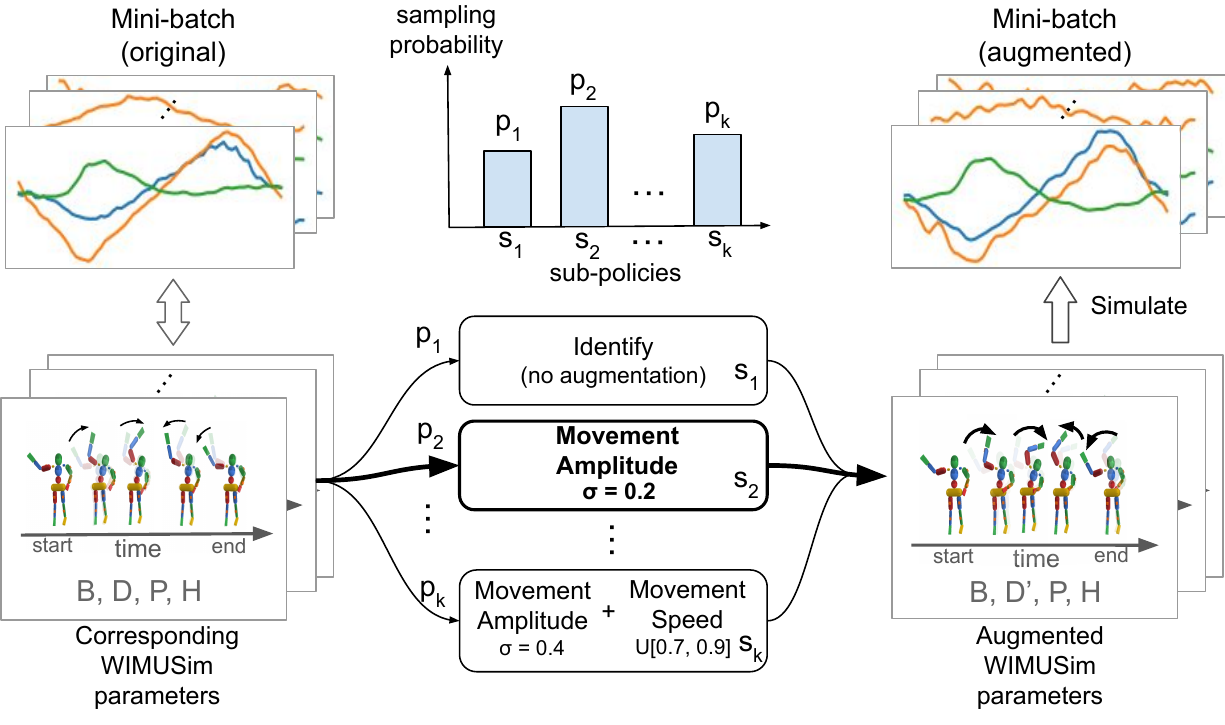}}
  \caption{Data augmentation pipeline during model training. Each mini-batch is augmented through a selected sub-policy $s_i$, sampled from a sub-policy set \{$s_1, s_2, ..., s_k$\} with probability $p_i$. Each sub-policy specifies one or more data augmentation methods and their parameters (see Table~\ref{tab:augmentation-params}). The modified WIMUSim parameters are then used to simulate augmented virtual IMU data.} 
\label{fig:da-process}
\end{figure}

For each dataset, WIMUSim parameters (\(B, D, P, H\)) were identified using real IMU data and corresponding motion capture or video-based pose estimation. For more details of the parameter identification process, refer to~\cite{oishi7wimusim}.
These identified parameters were then adjusted to introduce variations in virtual IMU data. As illustrated in Fig.~\ref{fig:da-process}, we applied data augmentation online, without generating a separate augmented dataset in advance. For each mini-batch, we sampled a sub-policy \(s_i\) from a predefined set \(\{s_1, s_2, ..., s_k\}\), each associated with a probability \(p_i\). Each sub-policy specified one or more augmentation methods and their parameter settings based on the experimental configurations; described later in Sections~\ref{sec:individual-comparison} and \ref{sec:reduce-data}). This allowed the model to encounter diverse and dynamically generated variations across epochs without significantly increasing storage requirements.  
In STDA, augmentations were applied directly to all IMU channels (accelerometer and gyroscope), whereas in PPDA, augmentations were applied by modifying WIMUSim parameters.

\begin{table}[h]
\scriptsize
\begin{threeparttable}
\caption{Parameter configurations for STDA and PPDA. See Table~\ref{tab:trans-aug-summary} for the definitions of STDAs and their parameters.}
\label{tab:augmentation-params}
\centering
\begin{tabular}{cp{5.8cm}}
\toprule
\multicolumn{1}{l}{\textbf{Method}}                                             & \textbf{Description and Parameter Configuration}   \\
\midrule
\begin{tabular}[c]{@{}c@{}}Movement Amplitude\\ (Magnitude Scaling)\end{tabular} & \begin{tabular}[c]{@{}p{5.8cm}@{}}
    STDA: Magnitude scaling with sample scaling factor \( \alpha \sim \mathcal{N}(1, \sigma^2) \), \( \sigma \in \{0.1, 0.2, 0.4, 0.6\} \).\\
    PPDA: Magnitude scaling, with the same sampling parameters as STDA, applied to \( D \) in WIMUSim.\end{tabular}                                         \\ \midrule
\begin{tabular}[c]{@{}c@{}}Movement Amplitude\\ (Magnitude Warping)\end{tabular} & \begin{tabular}[c]{@{}p{5.8cm}@{}}
    STDA: Magnitude warping with time-varying scale factor \( \boldsymbol{\alpha} \), generated with \( \sigma \in \{0.2, 0.4\} \) and \( K \in \{2, 4\} \).\\
    PPDA: Magnitude warping, with the same sampling parameters as STDA, applied to \( D \) in WIMUSim.\end{tabular}                                      \\ \midrule

\begin{tabular}[c]{@{}c@{}}Movement Speed\\ (Time Scaling)\end{tabular}     & \begin{tabular}[c]{@{}p{5.8cm}@{}}
    STDA: Time scaling with factor \( \beta \sim \mathcal{U}[\cdot] \) from ranges: \([0.7, 0.9]\), \([1.1, 1.3]\), \([0.75, 1.5]\), and \([0.5, 2.0]\). \\
    PPDA:  Time scaling, with the same sampling ranges as STDA, applied to $D$ in WIMUSim.\end{tabular}                         \\ \midrule
\begin{tabular}[c]{@{}c@{}}Movement Speed\\ (Time Warping)\end{tabular}     & \begin{tabular}[c]{@{}p{5.8cm}@{}}
    STDA: Time warping with \( K \in \{2, 4\} \), and max\_speed\_ratio\textsuperscript{*} \( \in \{1.5, 2.0\} \).\\
    PPDA: Time warping, with the same sampling parameters as STDA, applied to \( D \) in WIMUSim. \end{tabular}                                     \\ \midrule
Sensor Placement                                                       & \begin{tabular}[c]{@{}p{5.8cm}@{}}
    STDA: Rotation using matrices with angles sampled independently from \( \mathcal{U}[-180^\circ, 180^\circ] \) for each axis. \\
    PPDA: Sensor placement variation applied to \( P \) in WIMUSim, with per-axis orientation offsets sampled from \( \mathcal{U}[-25^\circ, 25^\circ] \), and dataset-specific modifications (see text).\end{tabular}     
    \\ \midrule
\begin{tabular}[c]{@{}c@{}}Hardware-related\\Effects\end{tabular}     & \begin{tabular}[c]{@{}p{5.8cm}@{}}
    STDA: Gaussian noise sampled from \(N(0, \sigma^2)\) (\(\sigma \in \{0.05, 0.1, 0.15, 0.2\}\)) added independently to each channel at every time step.
    \\
    PPDA: Same \( \sigma \) options for noise, with an additional constant bias term sampled per axis from \( \mathcal{U}(-1.0, 1.0) \), specified via \( H \) in WIMUSim. \end{tabular} \\
\bottomrule
\end{tabular}
\begin{tablenotes}
\footnotesize
\item $^{*}$Time warping in this study follows the \textit{tsaug} implementation~\cite{tsaug}, where the extent of deformation in the warping function $w(t)$ is constrained by \textit{max\_speed\_ratio}. For instance, setting \textit{max\_speed\_ratio} to 1.5 allows for up to a 50\% increase or decrease in the speed of segments defined by $K$ knots.
\end{tablenotes}
\end{threeparttable}
\end{table}

Table~\ref{tab:augmentation-params} presents the six data augmentation methods used in our experiments, categorized into four types: movement amplitude, movement speed, sensor placement, and hardware-related effects. Each method has both STDA and PPDA settings for comparison. While STDA directly transforms IMU signals using predefined mathematical operations (as summarized in TABLE \ref{tab:trans-aug-summary}), PPDA modifies WIMUSim parameters to introduce physically plausible variations.

For movement amplitude (scaling/warping) and movement speed (scaling/warping), STDA applies magnitude or temporal transformations directly to IMU signals, whereas PPDA modifies the corresponding $D$ parameters in WIMUSim to introduce realistic variations.

For sensor placement, STDA applies random 3D rotations to IMU signals within \([-180^\circ, 180^\circ]\) per axis, without accounting for sensor orientation constraints, as commonly done in the literature~\cite{um2017data, ohashi2017augmenting, mohammad2023enhanced}. In contrast, PPDA modifies the orientation component of \( P \) in WIMUSim by applying a fixed offset, sampled once per simulation run, from a default per-axis range of \( \mathcal{U}[-25^\circ, 25^\circ] \). Although this range is not tailored to specific devices or locations, it was conservatively chosen to avoid implausible configurations while capturing realistic variation due to body shape or device fitting. Additional dataset-specific adjustments are introduced where appropriate: In the REALDISP (sensor-displacement) scenario, to account for larger sensor misalignment in the test data, we additionally include a larger rotational misalignment sampled from \(\mathcal{U}[-90^\circ, 90^\circ]\) around the x-axis (aligned with the forearm), as well as discrete \(180^\circ\) flips along each axis to simulate reversed sensor placements.  In REALWORLD and MM-Fit, we also interchange \( P \) parameters among subjects to introduce inter-subject sensor placement variability, which involves both positional shift and orientation variations.

For hardware-related effects, PPDA specifies \( \sigma \) for noise and a constant bias term \( b \), independently sampled for each axis once per simulation run from \( \mathcal{U}[-1.0, 1.0] \), via the \( H \) parameter in WIMUSim.

\subsection{Individual Comparison of Augmentation Methods}
\label{sec:individual-comparison}

\begin{figure*}[t]
  \centering
  \begin{subfigure}[b]{1.0\textwidth}
      \begin{subfigure}[b]{0.24\textwidth}
        \centering
        \includegraphics[width=\textwidth]{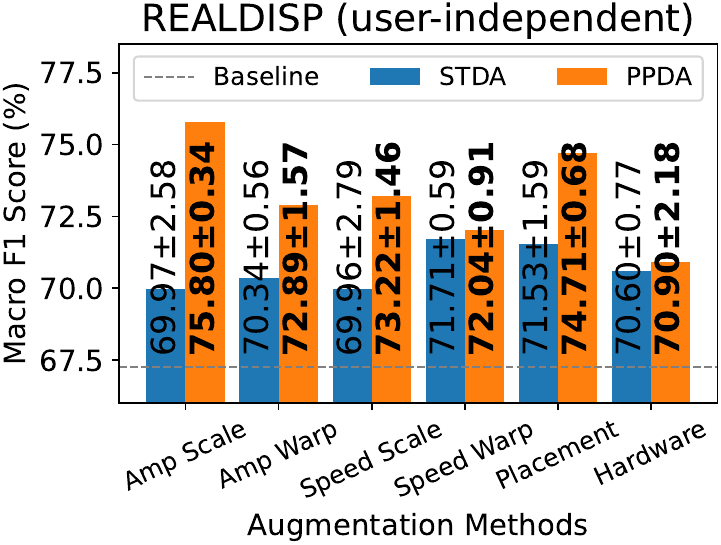}
      \end{subfigure}
      \begin{subfigure}[b]{0.24\textwidth}
        \centering
        \includegraphics[width=\textwidth]{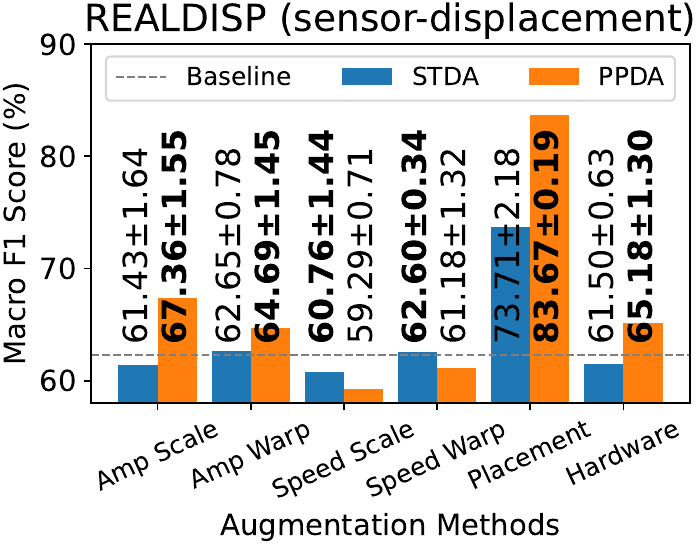}
      \end{subfigure}
      \begin{subfigure}[b]{0.24\textwidth}
        \centering
        \includegraphics[width=\textwidth]{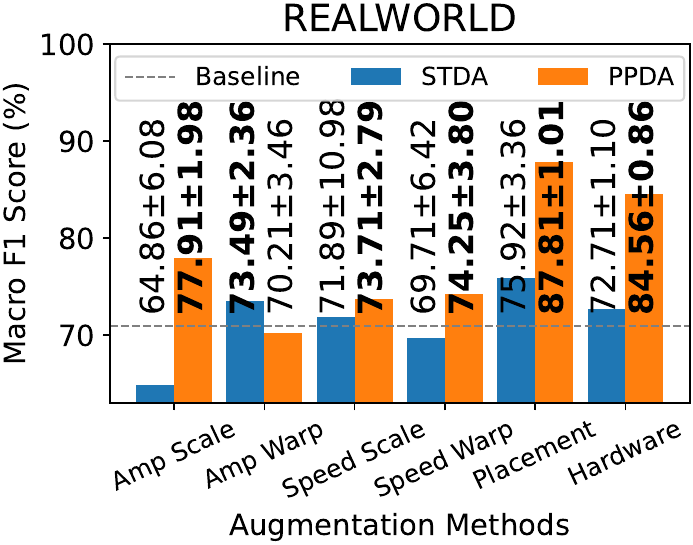}
      \end{subfigure}
      \begin{subfigure}[b]{0.24\textwidth}
        \centering
        \includegraphics[width=\textwidth]{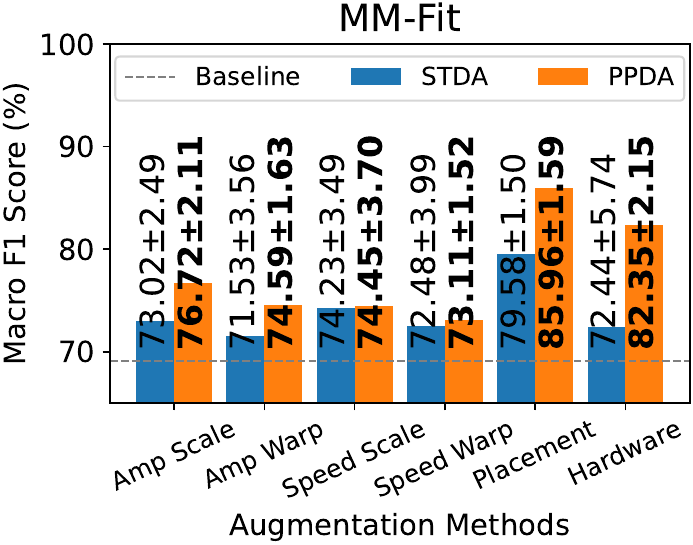}
      \end{subfigure}
      \caption{DeepConvLSTM}
  \end{subfigure}
  \hfill
  \begin{subfigure}[b]{1.0\textwidth}
      \begin{subfigure}[b]{0.24\textwidth}
        \centering
        \includegraphics[width=\textwidth]{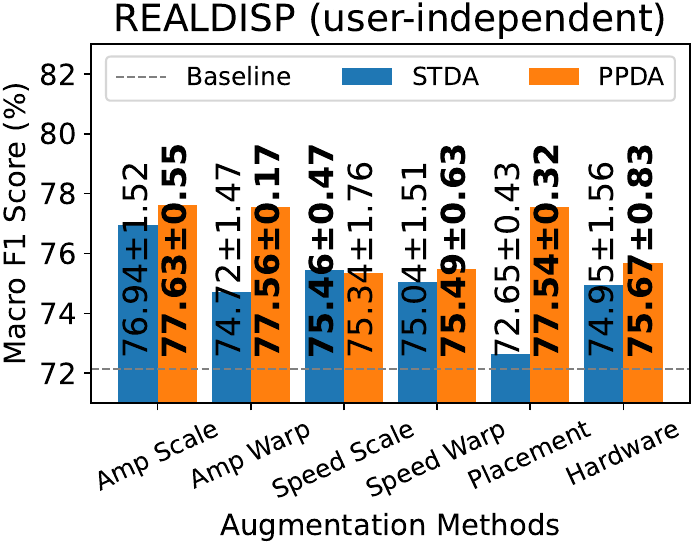}
      \end{subfigure}
      \begin{subfigure}[b]{0.24\textwidth}
        \centering
        \includegraphics[width=\textwidth]{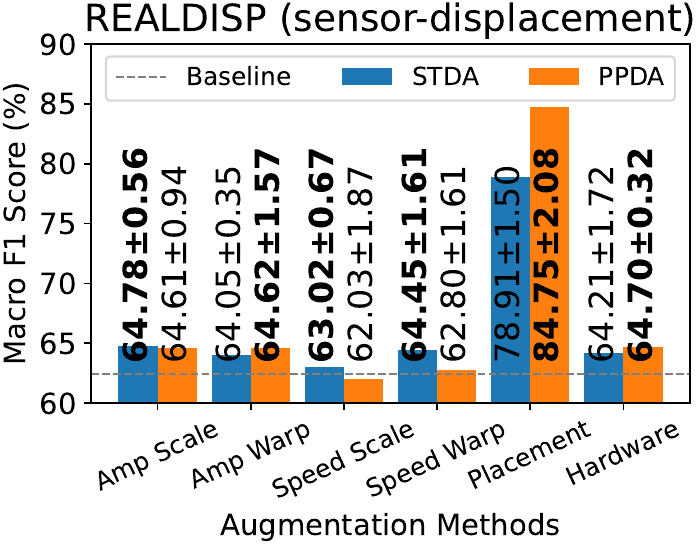}
      \end{subfigure}
      \begin{subfigure}[b]{0.24\textwidth}
        \centering
        \includegraphics[width=\textwidth]{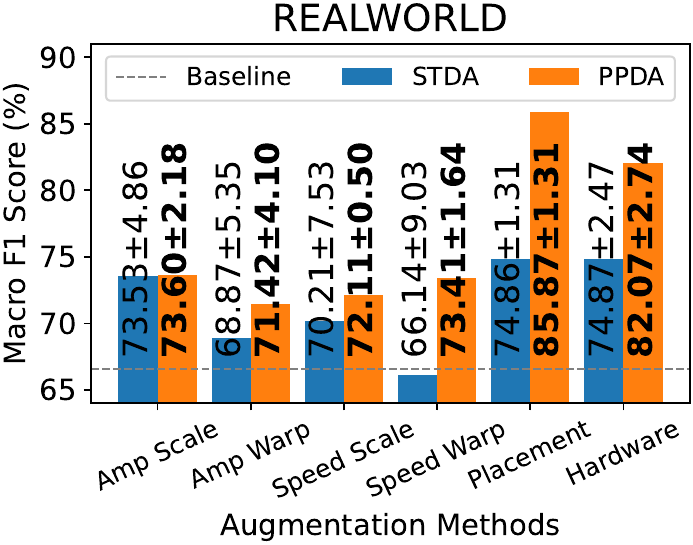}
      \end{subfigure}
      \begin{subfigure}[b]{0.24\textwidth}
        \centering
        \includegraphics[width=\textwidth]{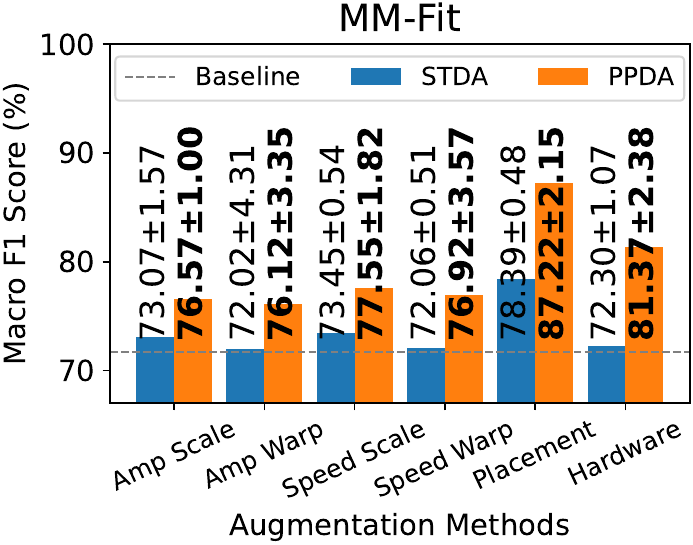}
      \end{subfigure}
      \caption{AttendAndDiscriminate}
    \end{subfigure}
    \caption{Comparison of macro F1 scores for STDA (blue) and PPDA (orange) methods across four experimental settings with two models: (a) DeepConvLSTM and (b) AttendAndDiscriminate. Each result reflects the best-performing parameter configuration for the corresponding augmentation method. The baseline represents performance using only real IMU data without augmentation. PPDA achieves comparable or significantly higher performance than its STDA counterpart in most cases, with an average improvement of 3.7 percentage points.}
    \label{fig:indiv-comparison}
\end{figure*}

To evaluate the benefits of physical plausibility in data augmentation, we compared each of the six PPDA methods with its STDA counterpart (listed in Table~\ref{tab:augmentation-params}) across four experimental settings. The REALDISP (user-independent), REALWORLD, and MM-Fit settings focused on generalization to unseen subjects under controlled or naturally varying sensor placements, while the REALDISP (sensor-displacement) setting specifically targeted robustness to sensor placement variation by training on ideal-placement data and testing on self-placement data.

In this experiment, we used a binary sub-policy setup, where each mini-batch was processed using either the identity sub-policy \(s_1\) (no augmentation) or an augmentation sub-policy \(s_2\) (applying a specific augmentation method with a specific parameter option). Both sub-policies were sampled with equal probability \(p_1 = p_2 = 0.5\). As a result, 50\% of the mini-batches were augmented while the remaining 50\% retained the original signals, balancing exposure to augmented data with retention of the original data distribution.

For augmentation methods that had multiple parameter options—specifically, movement amplitude (magnitude scaling, magnitude warping), movement speed (time scaling, time warping), and hardware-related effects—we tested each parameter option individually and reported the result for the best-performing parameter.

Fig.~\ref{fig:indiv-comparison} summarizes the macro F1 scores for all STDA and PPDA methods across the four experimental settings, using two deep learning models---DeepConvLSTM and AttendAndDiscriminate. The baseline reflects performance achieved using only real IMU data for training, without any augmentation.

Across the four experimental settings, PPDA methods generally outperformed their STDA counterparts, demonstrating the benefits of incorporating physically plausible variation. On average, PPDA yielded a 3.7 percentage point (pp) improvement in macro F1-score, with gains of up to 13 pp, highlighting the benefit of ensuring physical plausibility in data augmentation.

In the REALDISP (user-independent) setting, PPDA on movement amplitude (scaling) achieved the highest performance among all augmentation methods across both models, with the largest improvement observed in DeepConvLSTM. This is likely because different subjects performed exercise activities with varying styles and motion ranges, and simulating broader variations in movement amplitude helped the model generalize more effectively in this user-independent scenario.

In the REALDISP (sensor-displacement) setting, which evaluates robustness to sensor misplacement by training on ideal-placement data and testing on self-placement data, PPDA on sensor placement achieved the most significant improvement—likely because it explicitly modeled expected sensor orientation variations. In contrast, augmentations such as time scaling and time warping, both in STDA and PPDA, led to inconsistent or even degraded performance, as the variations they introduced did not directly address the sensor placement variability central to this scenario.

In the REALWORLD and MM-Fit settings—both of which evaluate generalization to unseen subjects using consumer-grade devices with naturally varying sensor placements—PPDA on sensor placement consistently yielded the highest performance gains across both models, followed by strong improvements from PPDA on hardware-related effects (noise and bias). These results highlight the importance of accounting for practical variability in sensor placement and sensor quality.

These findings highlight the benefits of physically plausible augmentation and underscore the importance of aligning augmentation strategies with the specific variations expected in the target environment.

\subsection{Impact on Reducing Data Collection Needs}
\label{sec:reduce-data}

To investigate whether PPDA can reduce the need for extensive data collection, we evaluated its effectiveness under limited labeled data conditions. Specifically, we assessed how well models generalize to unseen subjects when trained on datasets with limited inter-subject variation. In this setting, we combined multiple augmentation methods to introduce broader variation during model training. We compared PPDA with STDA to determine whether incorporating physical plausibility offers additional benefits in low-data regimes.

We conducted this experiment using the REALDISP (user-independent), REALWORLD, and MM-Fit settings, focusing on generalization to unseen subjects. For each dataset, we varied the number of subjects used for training data---from 1 to 10 for REALDISP and REALWORLD, and from 1 to 6 for MM-Fit---while keeping the validation and test sets fixed.

To introduce greater diversity during training, we combined multiple augmentation methods drawn from the four categories defined in Table~\ref{tab:augmentation-params}: movement amplitude, movement speed, sensor placement, and hardware-related effects. Each sub-policy \( s_i \) was constructed by choosing either the identity function (i.e. no augmentation) or one augmentation method from each of the four categories. For movement amplitude and movement speed, we included both scaling and warping augmentations, each with four parameter options. Sensor placement used one configuration, and hardware-related effects included four options. Including the option of the identity function in each category, this resulted in \( 9 \times 9 \times 2 \times 5 = 810 \) possible sub-policies $\{s_1, s_2, ..., s_{810}\}$ with initial sampling probabilities $p_1=p_2=...=p_{810}=\frac{1}{810}$.

Unlike the previous experiment, where augmentation was applied to only 50\% of mini-batches, here one of the 810 sub-policies was sampled for each mini-batch. One of these sub-policies was an identity (i.e., no-augmentation), while the others applied various combinations of augmentation methods. While some sub-policies may contribute positively to model generalization, others may have little effect or even negatively impact performance. Applying all sub-policies uniformly would dilute the impact of beneficial augmentations and limit the potential gains. To address this, we adopted a dynamic sub-policy sampling strategy inspired by AutoAugHAR~\cite{zhou2024autoaughar}, which allows the model to learn and prioritize augmentation combinations that are most effective during training. The sampling probabilities were optimized during training to maximize validation performance, allowing the model to prioritize more beneficial augmentation combinations.

\begin{figure*}[t]
  \centering'
  \begin{subfigure}[b]{0.9\textwidth}
      \begin{subfigure}[b]{0.33\textwidth}
        \centering
        \includegraphics[width=\textwidth]{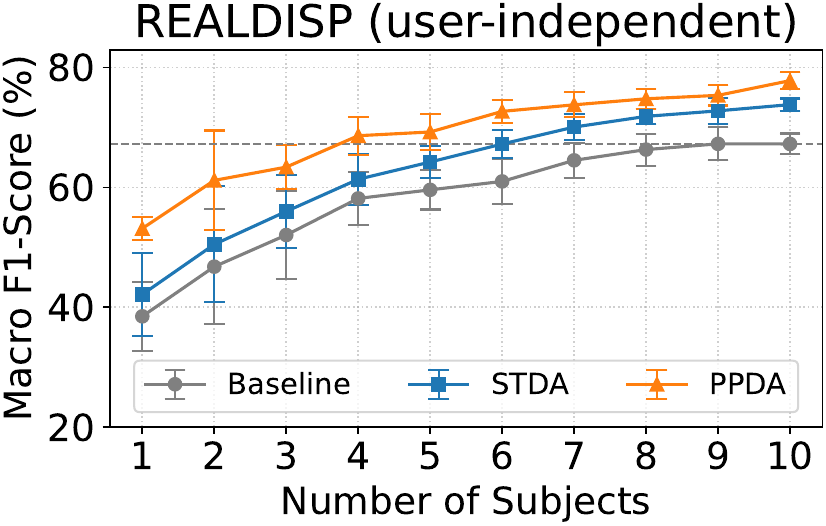}
      \end{subfigure}
      \begin{subfigure}[b]{0.33\textwidth}
        \centering
        \includegraphics[width=\textwidth]{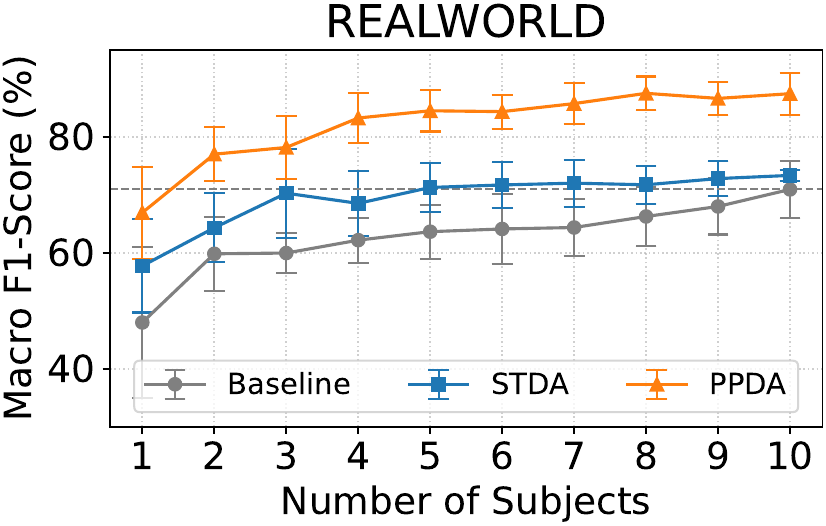}
      \end{subfigure}
      \begin{subfigure}[b]{0.33\textwidth}
        \centering
        \includegraphics[width=\textwidth]{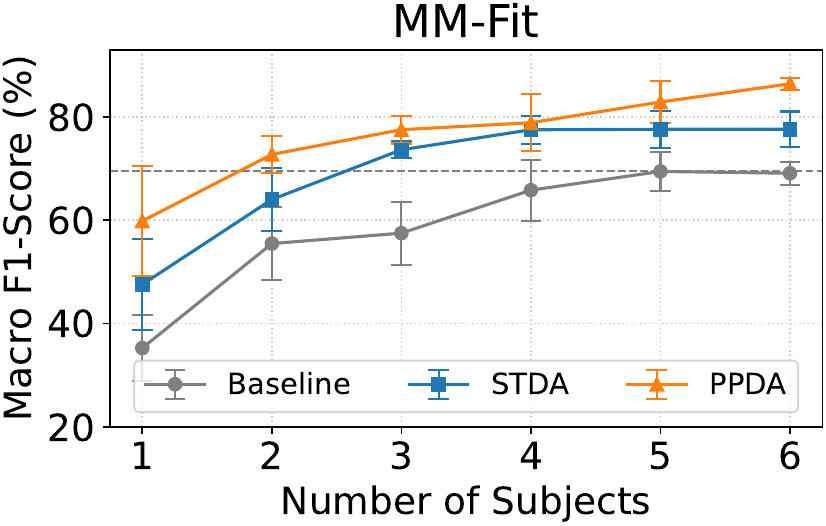}
      \end{subfigure}
      \caption{DeepConvLSTM}
  \end{subfigure}
  \hfill
  \begin{subfigure}[b]{0.9\textwidth}
      \begin{subfigure}[b]{0.33\textwidth}
        \centering
        \includegraphics[width=\textwidth]{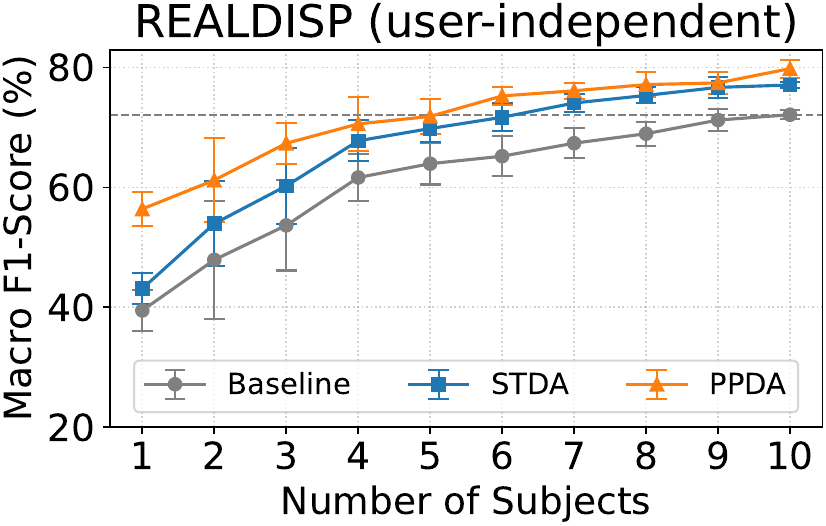}
      \end{subfigure}
      \begin{subfigure}[b]{0.33\textwidth}
        \centering
        \includegraphics[width=\textwidth]{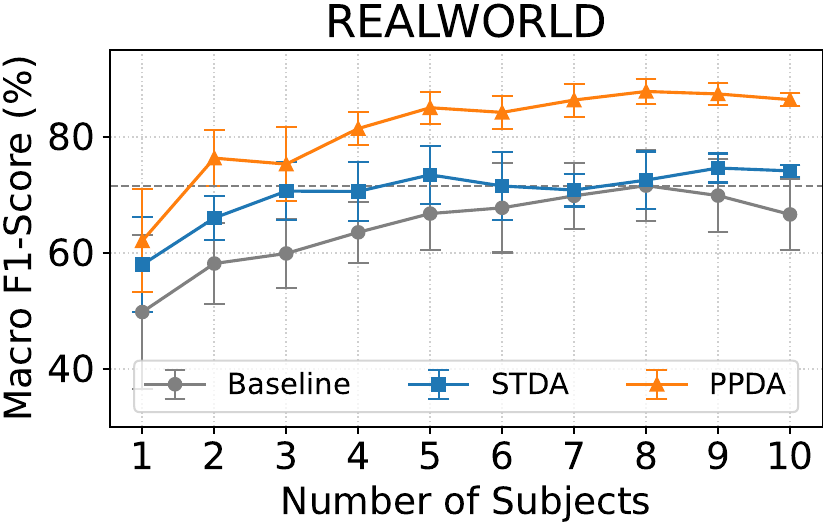}
      \end{subfigure}
      \begin{subfigure}[b]{0.33\textwidth}
        \centering
        \includegraphics[width=\textwidth]{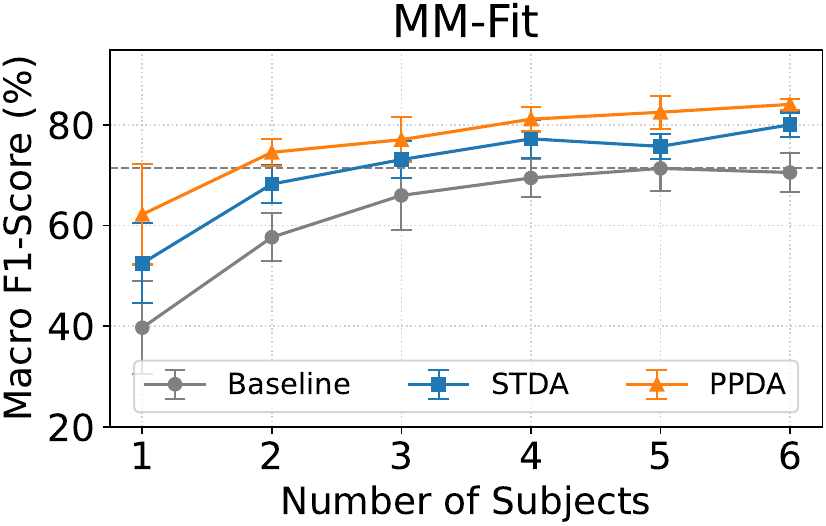}
      \end{subfigure}
      \caption{AttendAndDiscriminate}
    \end{subfigure}
    \caption{Macro F1 scores (\%) for (a) DeepConvLSTM and (b) AttendAndDiscriminate models across three datasets (REALDISP, REALWORLD, and MM-Fit) as the number of training subjects increases, comparing the baseline (no augmentation), STDA, and PPDA. The dashed horizontal line indicates the best baseline performance. PPDA consistently outperforms STDA and the baseline, with particularly large gains when the number of training subjects is small—demonstrating better generalization under limited inter-subject variation in training data.}
    \label{fig:n-sub-comparison}
\end{figure*}

Fig.~\ref{fig:n-sub-comparison} shows macro F1 scores for PPDA, STDA, and baseline (no augmentation) across increasing numbers of training subjects, evaluated on REALDISP, REALWORLD, and MM-Fit using DeepConvLSTM and AttendAndDiscriminate. To account for variation due to different training subject subsets, we repeated each experiment using 10 randomly sampled combinations of training subjects for each subset size. For example, for the 3-subject setting, we trained models on 10 randomly selected groups of 3 subjects and reported the average macro F1-score. For MM-Fit, we used all possible combinations when the number of combinations was fewer than 10 (e.g., 6 combinations for the 1-subject and 5-subject subsets), and randomly sampled 10 combinations otherwise. When using all subjects, we reported the average over three runs with different random seeds.

Across all datasets and models, PPDA consistently outperformed STDA, particularly when the number of training subjects was small. This demonstrates PPDA's effectiveness in enhancing model generalization in low-data regimes. For instance, with only one training subject, PPDA improved macro F1-scores by up to 13 pp in REALDISP, 9 pp in REALWORLD, and 12 pp in MM-Fit compared to STDA.

PPDA also reduced the number of training subjects required to reach the best baseline performance. In REALDISP, PPDA reached the best baseline performance with only 4–6 subjects, whereas STDA required 6–7. In REALWORLD and MM-Fit, PPDA achieved higher performance than the baseline with as few as 2 subjects, compared to 5 and 3 subjects needed for STDA, respectively. This corresponds to a reduction in the number of training subjects required by 40–80\% compared to the full baseline, and 14–60\% relative to STDA.

These results demonstrate that PPDA improves generalization to unseen subjects more effectively than STDA under limited supervision, supporting its potential to reduce data collection needs.



\section{Discussion}
Our results highlight the benefit of maintaining physical plausibility in data augmentation for wearable IMU-based HAR. By introducing physically plausible variations into the dataset through WIMUSim, PPDA enables deep learning models to generalize more effectively with limited training data.

One limitation of our current approach stems from its reliance on WIMUSim, which requires paired IMU and motion data to identify simulation parameters that ensure high-fidelity virtual IMU data. This dependency limits applicability in settings where such data are unavailable, e.g., the use of larger-scale video datasets where paired IMU data are unavailable. However, as pose estimation from RGB video and virtual IMU simulation techniques continue to improve, it may become feasible to estimate realistic simulation parameters without requiring paired IMU recordings—potentially expanding the applicability of our approach to broader scenarios.
Nevertheless, our PPDA method remains highly relevant in scenarios where small amounts of paired data can be collected—such as when developing HAR models for new wearable devices, where sensor data are typically captured alongside video, or in domains where data collection is especially costly or sensitive, such as clinical studies or research involving vulnerable populations.

Our PPDA methods have room for further refinement. While they demonstrated clear performance gains compared to STDAs, they do not yet fully leverage the flexibility offered by physics-based simulation. In our approach, we used simple yet physically plausible parameter modifications within WIMUSim, such as applying magnitude scaling or time warping to the $D$ parameters to simulate variations in movement amplitude or speed. However, these modifications were rather generic and not tailored to the specific characteristics of different activity types. For example, effective adjustments for dynamic exercises such as running or jumping may differ from those suited to more static activities like sitting or standing. Future work could explore activity-type-specific augmentation strategies, incorporating domain knowledge to create simulations better aligned with the unique movement dynamics of different activities.

\section{Conclusion}
In this study, we introduced and systematically characterized Physically Plausible Data Augmentation (PPDA) for wearable IMU-based HAR, leveraging physics simulation to introduce realistic variations across four categories: movement amplitude, movement speed, sensor placement, and hardware-related effects. To the best of our knowledge, this is the first study to systematically examine the role of physical plausibility in data augmentation for sensor-based HAR. We validated our approach across multiple public datasets covering 8 to 34 activity classes from daily activities and fitness workouts, using state-of-the-art deep learning models.

We demonstrated that maintaining physical plausibility in data augmentation significantly improves model generalization and reduces reliance on large-scale real-world data collection. Compared to Signal Transformation-based Data Augmentation (STDA) methods, PPDA achieved a notable improvement in model performance, with an average increase of 3.7 percentage points in macro F1-score in method-wise comparison (Section \ref{sec:individual-comparison}) and reductions in the number of required training subjects by up to 60\% (Section \ref{sec:reduce-data}).

Looking ahead, self-supervised learning also relies heavily on data augmentation, particularly contrastive learning techniques, which generate multiple views of the same data to learn meaningful representations without requiring labeled datasets~\cite{logacjov2024self}. PPDA presents a promising direction for self-supervised HAR, where physically meaningful augmentations may enhance feature learning and further reduce reliance on labeled datasets.

In conclusion, our findings position PPDA as an effective approach for improving the generalization of sensor-based HAR models, mitigating the challenges of data scarcity and the need for extensive data collection. While our results demonstrate the benefit of maintaining physical plausibility in data augmentation, the potential of physics-based simulation has not yet been fully leveraged. Further improvements in virtual IMU data fidelity, as well as incorporating more activity type-specific modifications, could further enhance HAR model performance and broaden the applicability of PPDA across a wider range of scenarios. To support further research and advancements in physics simulation-based data augmentation, we will make our PPDA implementation publicly available upon acceptance.


%





\ifCLASSOPTIONcaptionsoff
  \newpage
\fi



%
\bibliographystyle{IEEEtran}
\bibliography{bibtex/bib/main}

%

\begin{IEEEbiography}[{\includegraphics[width=1in,height=1.25in,clip,keepaspectratio]{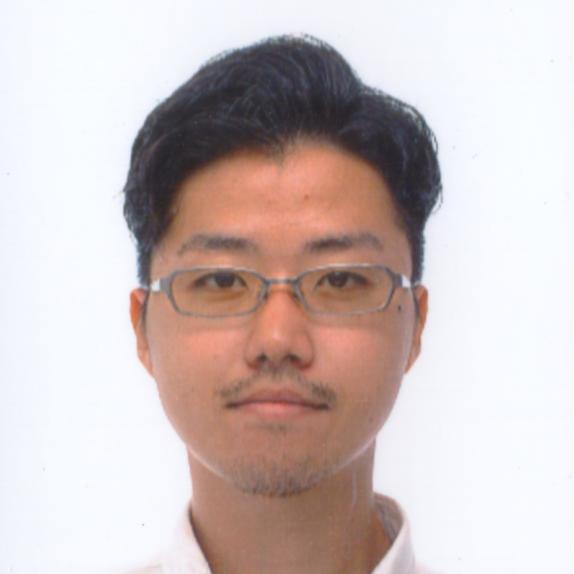}}]{Nobuyuki Oishi} is a PhD student at the Wearable Technologies Lab, University of Sussex, UK. His research focuses on leveraging physics simulations for wearable IMU-based Human Activity Recognition (HAR) to reduce the need for manual data collection and annotation, facilitating the development of HAR models that generalize across real-world variations, such as different sensor placements and subject characteristics. He received his B.Eng. and M.Eng. degrees from UEC Tokyo, Japan.
\end{IEEEbiography}
\begin{IEEEbiography}[{\includegraphics[width=1in,height=1.25in,clip,keepaspectratio]{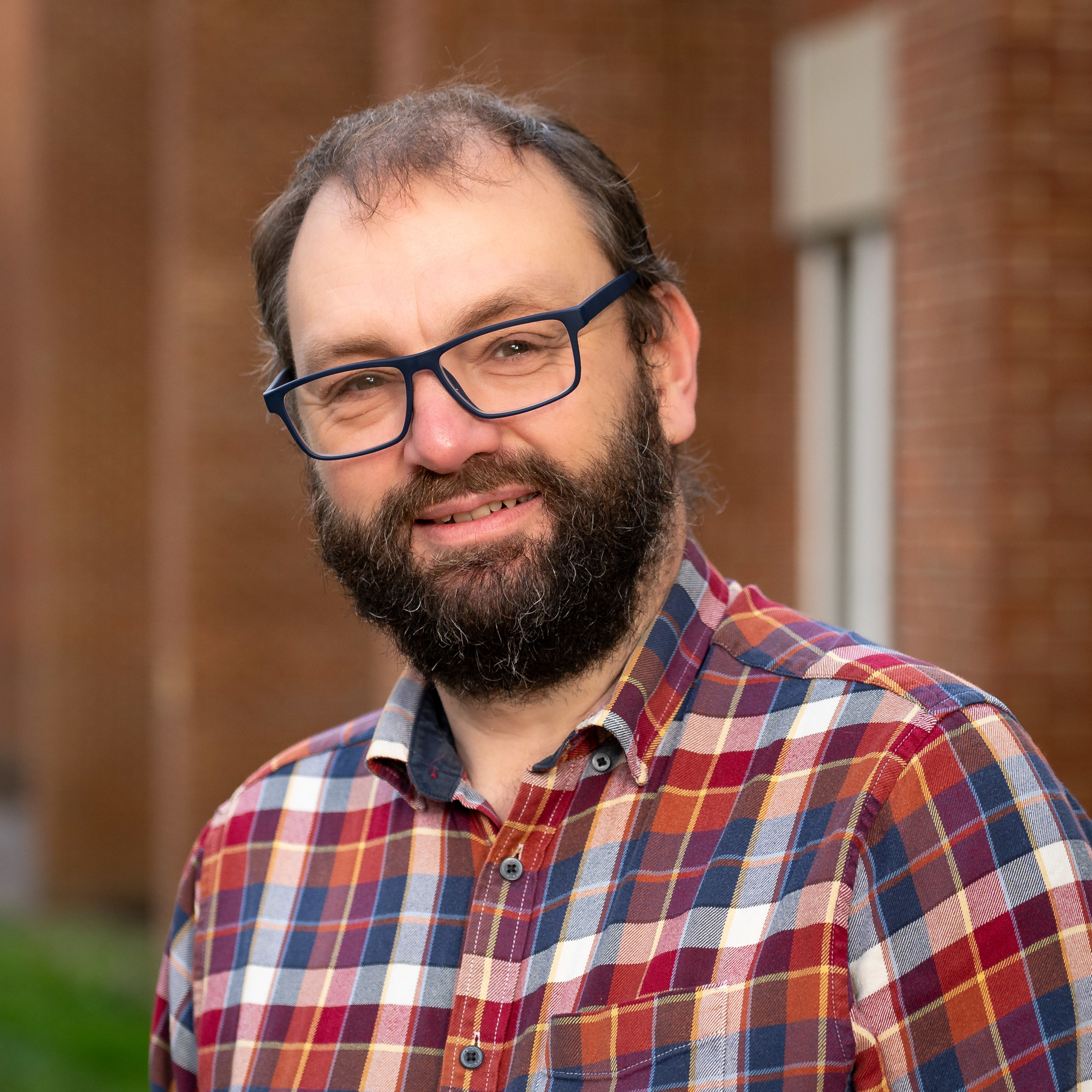}}]{Philip Birch} is an Associate Professor at the University of Sussex. His research interests include different imaging technologies, sensing, and computer vision. Application of this include the understanding of human activities using wearables devices. He also investigates using computer vision for multi-camera tracking of people or objects using edge based processing, and using this to detect unusual events. Dr Birch graduated from the University of Durham, UK with a PhD in 1999.
\end{IEEEbiography}
\begin{IEEEbiography}[{\includegraphics[width=1in,height=1.25in,clip,keepaspectratio]{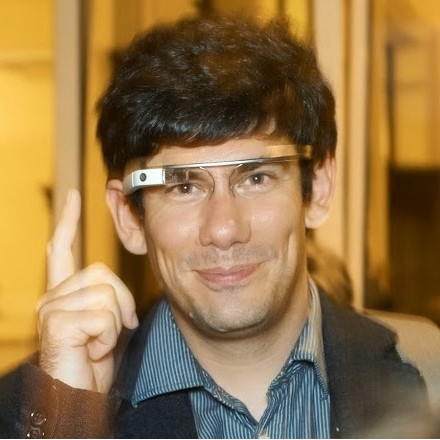}}]{Daniel Roggen} is a Professor in Wearable Technologies at the University of Sussex, UK, and research scientist at Google. His research focuses on computer behavior analytics in wearable and mobile computing: the art and science of understanding human behavior by the right combination of sensors, devices and AI techniques, and translating this into applications in HCI, sports, industry or health. His team published multimodal benchmark datasets and organized several ML challenges in human activity recognition. Prof. Roggen graduated from EPFL, Switzerland with a M.S. degree in microengineering in 2001 and Ph.D. degree in bio-inspired robotics in 2005.
\end{IEEEbiography}
\begin{IEEEbiography}[{\includegraphics[width=1in,height=1.25in,clip,keepaspectratio]{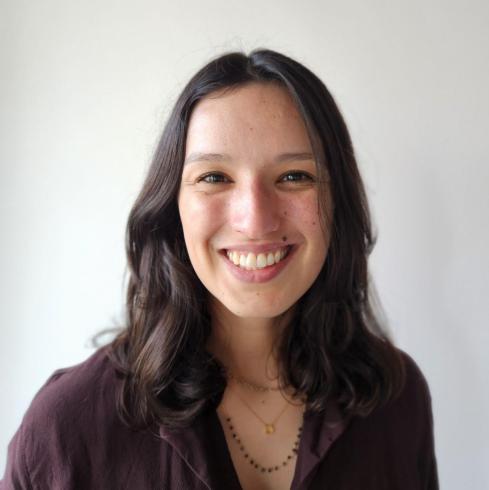}}]{Paula Lago} is an Assistant Professor at Concordia University, Montreal, Canada. Her research interests lie at the intersection of Pervasive Healthcare, Machine learning, and data analysis, with a focus on understanding the relationships between human behavior and health outcomes. Previously, she worked as a Postdoctoral Researcher at the Kyushu Institute of Technology in Japan, where she explored the application of wearable sensors for activity recognition and developed approaches to integrating additional sensor data during the training phase of machine learning models.
\end{IEEEbiography}




\end{document}